%% file: acl_latex.tex
\documentclass[11pt]{article}

\usepackage{acl}

\usepackage{times}
\usepackage{latexsym}

\usepackage[T1]{fontenc}

\usepackage[utf8]{inputenc}

\usepackage{microtype}

\usepackage{inconsolata}

\usepackage{graphicx}

\usepackage{xspace}
\newcommand{\ours}{\textsc{GrocLM}\xspace}
\usepackage{tcolorbox}
\usepackage{amsmath}
\newtcolorbox{mybox}{colback=red!5!white,colframe=red!75!black}
\usepackage{tikz}
\usepackage{pgfplots}
\pgfplotsset{compat=1.18}

\title{GrocLM: Grocery Category Recommendation in E-Commerce with Large Language Models}

\author{%
\textbf{Yuan Zhong}$^1$\;\;\; \textbf{Chuanwei Ruan}$^2$\;\;\; \textbf{Moein Hasani}$^2$ \\
\textbf{Tejaswi Tenneti}$^2$\;\;\; \textbf{Haixun Wang}$^3$\;\;\; \textbf{Fenglong Ma}$^1$\thanks{Corresponding Author.} \\
$^1$The Pennsylvania State University, USA \\
$^2$Instacart, USA \\
$^3$Evenup, USA \\
\texttt{$^1$\{yfz5556, fenglong\}@psu.edu} \\
\texttt{$^2$\{chuanwei.ruan, moein.hasani, tejaswi.tenneti\}@instacart.com} \\
\texttt{$^3$haixun.wang@evenup.ai}
}

\begin{document}
\maketitle
\begin{abstract}
The rapid growth of online grocery shopping requires recommendation systems that capture cyclical purchasing behavior and diverse user intents. Traditional item-level methods face scalability and accuracy challenges, motivating category-level recommendation as a more structured and practical alternative. We present \ours, a fine-tuned language model for grocery category recommendation in a real-world production environment. \ours\ employs a two-stage LoRA-based training strategy to encode cyclical purchasing patterns directly into model parameters, enabling more effective utilization of rebuying signals compared to prompt-based conditioning. To ensure valid and controllable outputs, we further introduce a trie-based constrained decoding mechanism over a predefined category space. Experiments on both proprietary production data and a public benchmark demonstrate that \ours\ consistently outperforms strong baselines. In a live production restocking task, \ours\ achieves a 7.5\% relative improvement in cart-adds per impression, while maintaining efficient inference by generating all categories jointly. These results highlight the effectiveness and practicality of integrating large language models into structured recommendation systems.
\end{abstract}

\input{inputs/1-intro}
\input{inputs/2-relatedWork}
\input{inputs/3-method}

\input{inputs/4-experiment}

\input{inputs/5-conclusion}

\newpage
\section{Ethical Considerations}

This work is based on a proprietary large-scale grocery dataset derived from real-world user interactions. All data used in this study were anonymized and aggregated in accordance with internal data governance policies. No personally identifiable information was accessed or utilized during model development or evaluation. The system operates at the product category level and does not infer or generate sensitive personal attributes.

As a deployed recommendation system, \ours may influence product exposure and customer purchasing decisions. To mitigate potential risks, we incorporate constrained generation mechanisms that restrict outputs to a predefined category vocabulary, reducing the risk of unsafe or inappropriate predictions. We also conduct offline evaluation and controlled production testing prior to deployment. Continuous monitoring is performed to detect performance degradation, unintended biases, or distribution shifts. Human oversight remains part of the deployment workflow to ensure system reliability and alignment with business and user experience goals.

\section{Broader Impact}

Category-level recommendation systems can improve user experience by reducing search friction and helping customers efficiently discover relevant products. For large-scale online grocery platforms, improved recommendation accuracy may also enhance operational efficiency and reduce cognitive load for users navigating extensive catalogs.

However, recommendation systems may also amplify existing popularity biases or disproportionately favor certain product groups. While \ours does not model sensitive demographic attributes, systematic biases in historical purchasing data may influence predictions. Future work includes more formal bias auditing and fairness-aware optimization strategies to ensure balanced exposure across product categories. By emphasizing constrained generation and deployment safeguards, we aim to promote responsible and trustworthy application of large language models in real-world commerce environments.

\bibliography{custom}

\appendix
\input{inputs/7-supp}



\end{document}

%% file: inputs/1-intro.tex
\section{Introduction}

At a large-scale online grocery e-commerce platform, we serve millions of customers daily across thousands of retailers. Unlike traditional e-commerce, grocery shopping is mission-driven and highly repetitive: users frequently repurchase household staples and build baskets spanning multiple complementary categories. This recurring nature requires recommendation systems that model structured shopping habits and replenishment patterns, not just isolated item preferences.

Our platform adopts a hybrid strategy combining item-level and carousel-level models. While item-level recommendation remains essential, category-level recommendation provides a more scalable and business-aligned solution. On the storefront, users interact with category-based carousels that structure product exposure and ensure inventory availability. In practice, category recommendation is typically implemented in a \textit{top-down} manner: predicting categories first and then populating them with items, because grocery scenarios require strict diversification and availability constraints that bottom-up aggregation struggles to satisfy.

Category-level modeling is also better aligned with grocery behavior. Repurchasing patterns often emerge at the category level rather than the item level: while specific products (e.g., \textit{Gala Apple} vs. \textit{Fuji Apple}) may vary, demand for the broader \textit{Apple} category remains stable. Moreover, grocery baskets are typically large and span complementary categories, requiring models to understand sequential intent and evolving shopping missions.

However, designing effective grocery category recommendation models is non-trivial. Existing LLM-based generative recommendation approaches~\cite{rajput2024recommender, si2023generative, geng2022recommendation, zheng2024adapting, tan2024towards, wang2024enhanced} are primarily developed for item-level prediction and do not explicitly model cyclical repurchasing behavior. Semantic ID-based methods further introduce potential noise through intermediate ID mapping. Directly feeding purchase histories into LLMs is appealing, yet unconstrained generation produces free-form text that conflicts with the mutually exclusive and discrete nature of category labels. These challenges call for a tailored solution.

We propose \ours, a prompt-tuned \underline{\textbf{L}}anguage \underline{\textbf{M}}odel for \underline{\textbf{Groc}}ery category recommendation. Built on a LLAMA 3~\cite{dubey2024llama} backbone with Low-Rank Adapters (LoRA)~\cite{hu2021lora}, \ours adopts a two-stage fine-tuning strategy.

In Stage 1, we model \textbf{category-level repurchasing behavior} using pre-computed rebuying statistics, enabling the model to capture high-level cyclical patterns and improve robustness under sparse histories. In Stage 2, the model learns \textbf{implicit relationships among categories} from sequential purchasing data, augmented with user query context to better reflect immediate intent. To ensure valid and discrete outputs, we introduce a trie tree-based masking mechanism~\cite{de2020autoregressive} with beam search to constrain decoding to legitimate category labels.

The key contributions of this work are:
\begin{itemize}
\item We formalize grocery category recommendation as a practical task characterized by cyclical purchasing patterns, diverse user queries, and discrete output constraints.
\item We propose \ours, a two-stage prompt-driven LLM framework leveraging rebuying statistics, sequential behavior modeling, and trie-constrained decoding.
\item We validate \ours on both proprietary and public datasets, demonstrating significant improvements over strong baselines and measurable gains in production metrics.
\end{itemize}

\begin{figure}[t]
    \centering
    \vspace{0.2in}
    \includegraphics[width=1\columnwidth]{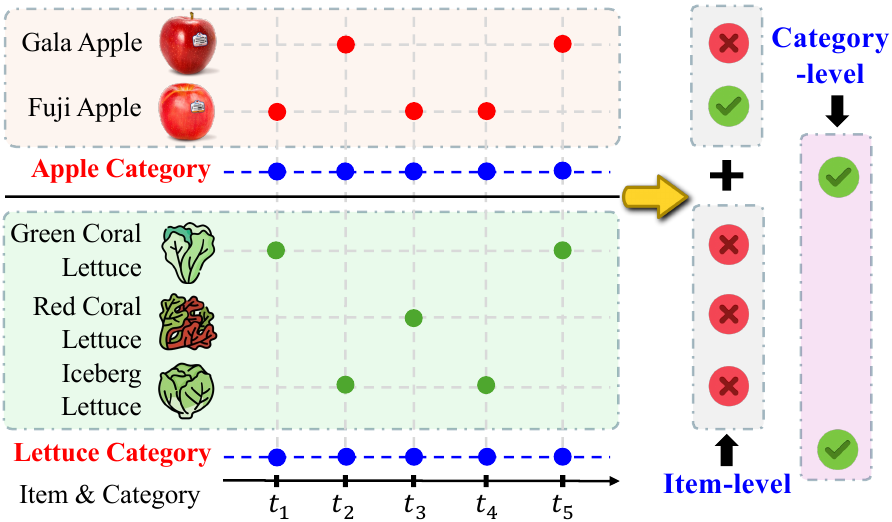}
    \vspace{-0.2in}
    \caption{Illustration of grocery category recommendation: aggregating items into categories reveals clear repurchasing patterns.}
    \label{fig:itemCate}
    \vspace{-0.15in}
\end{figure}

\begin{figure*}[t]
    \centering
\includegraphics[width=0.85\textwidth]{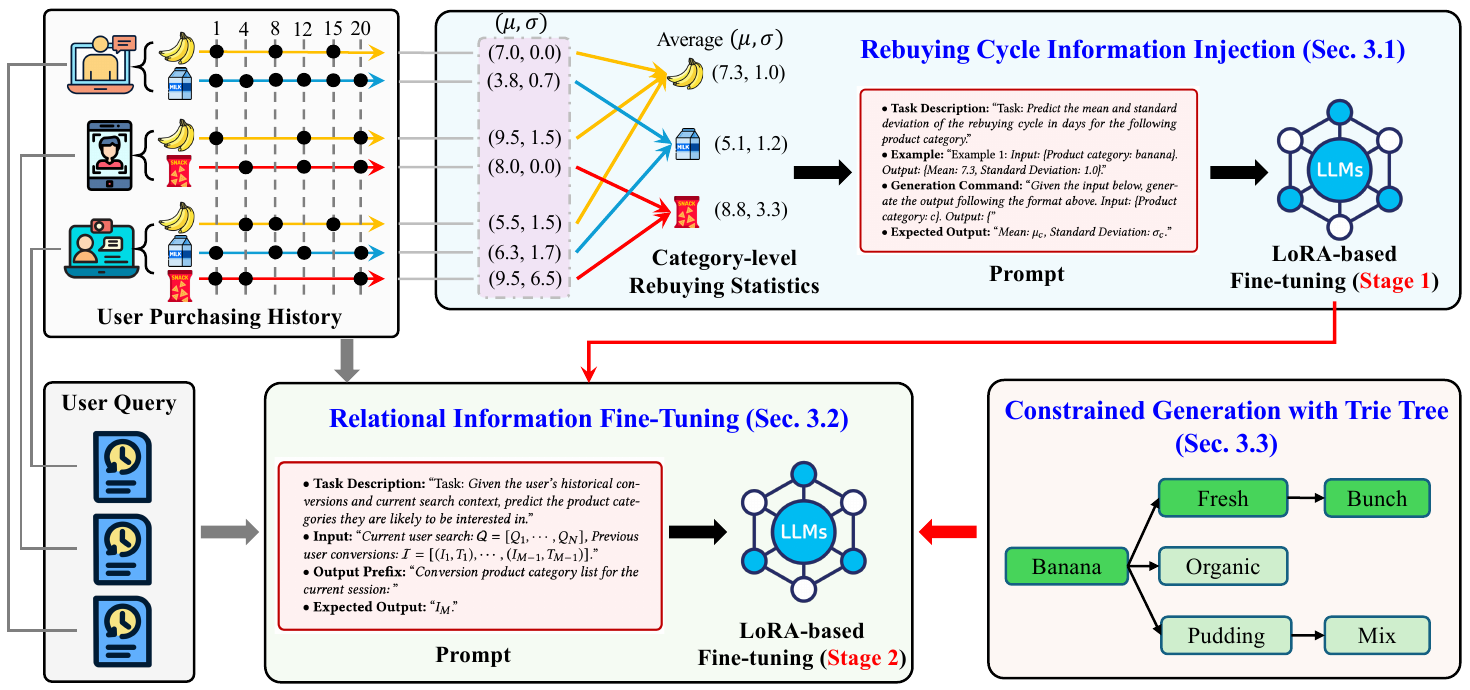}
    \vspace{-0.1in}
    \caption{Overview of the proposed \ours}
    \label{fig:mainModel}
\vspace{-0.05in}
\end{figure*}

%% file: inputs/2-relatedWork.tex
\section{Related Work}

\noindent \textbf{Generative Recommendation with LLMs.}
Recent advances in large language models (LLMs)~\cite{deldjoo2024review,xiao2022training,chen2024large,zhao2024recommender,wu2024survey,zhang2023chatgpt,wu2021empowering} have significantly influenced recommendation systems. Leveraging pre-trained encoders and decoders, these models support diverse outputs, including item identifiers~\cite{harte2023leveraging,mao2023unitrec,sanner2023large,sileo2022zero}, user ratings~\cite{bao2023tallrec,kang2023llms}, and text-based recommendations~\cite{li2020generate,li2023personalized,ni2019justifying,hada2021rexplug}. Their flexibility enables zero-shot recommendation~\cite{dai2023uncovering,kang2023llms,zhang2023chatgpt,liu2023chatgpt,sanner2023large,sileo2022zero} and domain adaptation through fine-tuning~\cite{bao2023tallrec,cui2022m6,hua2023index,geng2022recommendation,harte2023leveraging}. Generative recommender models extend these capabilities by directly generating item identifiers, reducing reliance on traditional ranking pipelines. Approaches incorporating semantic identifiers~\cite{geng2022recommendation,rajput2024recommender,tan2024towards}, collaborative filtering~\cite{zheng2024adapting,wang2024enhanced,khattab2020colbert,li2021causalbert,he2020lightgcn}, LLM-based architectures~\cite{bao2024real,kim2024large}, and autoencoder techniques~\cite{si2023generative} have streamlined recommendation workflows. Despite these advances, grocery recommendation remains underexplored. Off-the-shelf LLMs are pre-trained on general-domain corpora~\cite{deldjoo2024review}, making grocery data inherently out-of-distribution. Additionally, privacy constraints limit large-scale domain-specific pre-training. As a result, challenges such as cyclical purchasing behavior and diverse user queries are insufficiently addressed.

\noindent \textbf{Pre-LLM Retrieval and Recommendation Methods.}
Prior to LLM-based approaches, sparse retrieval methods such as vector space models~\cite{salton1962some,salton1975vector}, TF-IDF~\cite{aizawa2003information,ramos2003using,robertson2004understanding}, and inverted indices~\cite{zobel2006inverted,zobel1998inverted} dominated search and recommendation. While computationally efficient, these methods relied heavily on keyword matching and lacked semantic understanding. Learning-to-rank techniques~\cite{liu2009learning} improved relevance through supervised learning but remained constrained by sparse representations. Dense retrieval models~\cite{huang2013learning,guo2016deep,guo2019matchzoo,mitra2017neural,kang2018self} later introduced learned embeddings to capture semantic similarity. Architectures such as Two Tower models and self-attentive sequential networks better modeled user behavior and item relationships. The emergence of transformer-based models~\cite{vaswani2017attention} and BERT~\cite{devlin2018bert} further enhanced contextualized representation learning. Bi-encoder methods~\cite{karpukhin2020dense,qu2020rocketqa,xiong2020approximate,ni2021sentence,ni2021large,reimers2019sentence} and multi-representation models~\cite{humeau2019poly,luo2022improving,khattab2020colbert,hofstatter2022introducing} enabled efficient and fine-grained query-document matching. However, even dense retrieval and pre-trained embedding models~\cite{guo2022semantic,fan2022pre,yates2021pretrained} do not explicitly model domain-specific constraints such as cyclical repurchasing patterns and discrete category outputs in grocery recommendation.

%% file: inputs/3-method.tex
\section{Methodology}
This work aims to predict accurate grocery category recommendations based on users' historical interactions and current queries.

Let $\mathcal{I} = [(I_1, T_1), \cdots, (I_{M-1}, T_{M-1})]$ denote a historical interaction sequence of length $M-1$, where each interaction $I_m$ at time $T_m$ is a subset of categories, i.e., $I_m \subset \mathcal{C}$ and $\mathcal{C}$ is the set of all grocery categories. We use $\mathcal{Q} = [Q_1, \cdots, Q_N]$ to denote the current $N$ user queries.

Given $\mathcal{I}$, $\mathcal{Q}$, an LLM $\mathcal{F}$, and a set of prompt templates $\mathcal{P} = [P_1, \cdots, P_K]$, the goal is to predict the category set at time $M$, denoted as $I_M$. Since the task requires generating \textbf{discrete and mutually exclusive category labels}, we formulate it as autoregressive generation over category tokens:
\begin{equation}
p(I_M) =
\sum_{c \in I_M}
\prod_{l=1}^{L_c}
p(w_l \mid \mathcal{F}, \mathcal{I}, \mathcal{Q}, \mathcal{P}, w_{<l}).
\end{equation}

where a category $c = [w_1, \cdots, w_{L_c}]$ consists of $L_c$ tokens, and $w_{<l} = [w_1, \cdots, w_{l-1}]$ (or $\emptyset$ if $l=1$) denotes preceding tokens.

To address this formulation, we propose \ours, a generative retrieval framework based on prompt tuning and LLM fine-tuning (Figure~\ref{fig:mainModel}). The model includes three components: (1) Rebuying Cycle Information Injection, (2) Relational Information Fine-Tuning, and (3) Constrained Generation with a Trie Tree. We detail each module below.

\subsection{Rebuying Cycle Information Injection}

A naive strategy is to fine-tune the LLM $\mathcal{F}$ directly on user interaction history $\mathcal{I}$ and queries $\mathcal{Q}$. However, this overlooks explicit modeling of cyclical purchasing behavior, which is fundamental in grocery data. These recurring patterns provide an unconditional temporal prior over category demand, enabling the model to make reasonable predictions even with limited interaction history. Therefore, we first inject category-level rebuying statistics before modeling more context-dependent relational patterns.

\noindent\textbf{Category-level Rebuying Statistics Calculation.}
For each category $c \in \mathcal{C}$ and user $u \in \mathcal{U}$ with at least two conversions, we compute the time gaps between consecutive purchases. Let $\{T^{u,c}_1, \cdots, T^{u,c}_t\}$ denote the timestamps of purchases for category $c$. The time gaps are defined as $\Delta T_i^{u,c} = T_i^{u,c} - T_{i-1}^{u,c}$ for $i \ge 2$. The mean and standard deviation of these gaps for user $u$ are denoted as $\mu_{u,c}$ and $\sigma_{u,c}$.

Aggregating across users yields the category-level statistics:
\begin{equation}
    \mu_c = \frac{1}{|\mathcal{U}|}\sum_{u=1}^{|\mathcal{U}|} \mu_{u,c}, \;\;
    \sigma_c = \frac{1}{|\mathcal{U}|}\sum_{u=1}^{|\mathcal{U}|} \sigma_{u,c},
\end{equation}
where $|\mathcal{U}|$ is the number of users.

\noindent\textbf{Prompt-based Rebuying Pattern Injection with LoRA.}
To inject the rebuying statistics $\{c, \mu_c, \sigma_c\}$ into the model $\mathcal{F}$, we adopt a prompt-based LoRA~\cite{hu2021lora} fine-tuning approach. Specifically, we design the following prompt:

\begin{mybox}
$\bullet$ \textbf{Task Description:} ``Task: \textit{Predict the mean and standard deviation of the rebuying cycle in days for the following product category}.'' 

$\bullet$ \textbf{Example:} ``Example 1: \textit{Input: \{Product category: banana\}. Output: \{Mean: 7.3, Standard Deviation: 1.0\}}.'' 

$\bullet$ \textbf{Generation Command:} ``\textit{Given the input below, generate the output following the format above. Input: \{Product category: $c$\}. Output: \{}'' 

$\bullet$ \textbf{Expected Output:} ``\textit{Mean: $\mu_{c}$, Standard Deviation: $\sigma_{c}$.}'' 
\end{mybox}
Note that we have $|\mathcal{C}|$ categories in the training set, and the corresponding $|\mathcal{C}|$ category-level statistics $\{c, \mu_c, \sigma_c\}$ ($\forall c \in \mathcal{C}$) are used to fine-tune $\mathcal{F}$ with LoRA.

\subsection{Relational Information Fine-Tuning}

Although Stage 1 captures high-level cyclical patterns, it does not model personalized preferences or session-level intent. Aggregated rebuying statistics overlook user-specific behaviors and in-session relationships among multiple queries. To address this, we introduce Relational Information Fine-Tuning as the second training stage. Similar to Stage 1, this phase adopts LoRA-based prompt tuning, but focuses on sequential interaction history and current queries to model context-aware user behavior.

Specifically, we treat the user’s current search session as the \textbf{context}, and historical conversions as the \textbf{condition}. The context reflects short-term intent through queries (e.g., ``fruit'' or ``dairy''), while the condition captures long-term purchasing patterns from historical interactions. Our prompt is structured as follows:

\begin{mybox} 
$\bullet$ \textbf{Task Description:} ``Task: \textit{Given the user's historical conversions and current search context, predict the product categories they are likely to be interested in.}'' 

$\bullet$  \textbf{Input:} ``\textit{Current user search: $\mathcal{Q}= [Q_1,\cdots,Q_{N}]$, Previous user conversions: $\mathcal{I} = [(I_1, T_1),\cdots, (I_{M-1}, T_{M-1})]$.}'' 

$\bullet$  \textbf{Output Prefix:} ``\textit{Conversion product category list for the current session: }'' 

$\bullet$  \textbf{Expected Output:} ``$I_M$.'' 
\end{mybox}

A straightforward approach is to apply standard autoregressive training by treating $I_M$ as a sentence~\cite{geng2022recommendation,tan2024towards}. However, this is sub-optimal for two reasons. First, categories in $I_M$ are independent and mutually exclusive, differing from natural language tokens. Second, unconstrained generation requires post-hoc matching to valid categories, introducing errors. Even minor token deviations (e.g., ``\textit{snack}'' vs. ``\textit{snacks}'') lead to false negatives. Therefore, additional output constraints are necessary beyond LoRA fine-tuning alone.

\subsection{Constrained Generation with Trie Tree}\label{trie}

\begin{figure}[t]
    \centering
    \includegraphics[width=0.95\columnwidth]{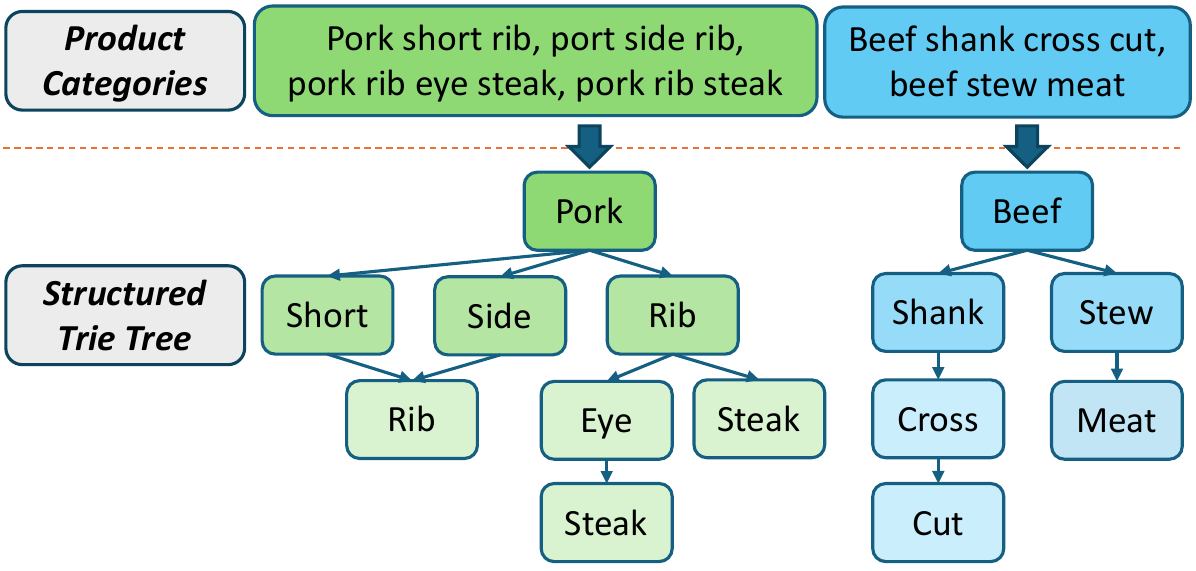}
    \vspace{-2mm}
    \caption{Illustration of trie tree.}
    \label{fig:trie}
    \vspace{-0.1in}
\end{figure}

To address the discrete and mutually exclusive nature of category outputs, we adopt a trie-based constrained decoding strategy inspired by~\cite{de2020autoregressive}. All valid product categories are tokenized and organized into a trie tree $\mathcal{T}$ (Figure~\ref{fig:trie}). Each node corresponds to a token, and its children define valid subsequent tokens in a category sequence.

During decoding, the next token $w_l$ is constrained to a valid set $\mathcal{V}_l$ defined by the trie. For example, given the prefix ``\textit{[Pork, Rib]}'', the valid candidates are ``\textit{[Eye, Steak]}''. This masking ensures that only valid category sequences can be generated. Let $\mathbf{x} = (\mathcal{F}, \mathcal{I}, \mathcal{Q}, \mathcal{P}, \mathcal{T})$ denote the full conditioning context. The constrained probability is defined as:

\begin{equation}
p(w_l) =
\begin{cases}
p(w_l \mid \mathbf{x}, w_{<l}) & \text{if } w_l \in \mathcal{V}_l, \\
0 & \text{otherwise}.
\end{cases}
\end{equation}

The training objective of \ours is therefore:
\begin{equation}
\min \mathcal{L}
= -\sum_{c \in I_M}
\sum_{l=1}^{L_c}
\log p(w_l \mid \mathbf{x}, w_{<l}).
\end{equation}

During inference, we apply beam search while enforcing trie-based constraints at each step. New categories can be incorporated by updating the trie without retraining the model.

%% file: inputs/4-experiment.tex
\section{Experiments}
In this section, we compare our model's performance on various metrics, datasets, and prediction tasks.

\subsection{Datasets}

We evaluate \ours\ on a proprietary large-scale grocery dataset, referred to as the \emph{Conversion} dataset. 
Statistics of the \emph{Conversion} dataset are shown in Table~\ref{table:dataset}.%
\footnote{In addition to the proprietary \emph{Conversion} dataset, we also conduct experiments on a public benchmark, the InstaCart Online Grocery Basket Analysis Dataset~\cite{instacart2021} (\url{https://www.kaggle.com/datasets/yasserh/instacart-online-grocery-basket-analysis-dataset}), referred to as the \emph{Aisle} dataset. Results on this dataset are reported in Appendix~\ref{exp2}.}

\textbf{Conversion Dataset.}
The \emph{Conversion} dataset is derived from a large online grocery e-commerce platform and contains user query and conversion histories. We extract search queries, timestamps, and associated product categories, retaining users with at least five conversions within a six-month window and at least ten total purchases. Each record represents a user’s query–conversion sequence in JSON format. We also construct a Rebuying Statistics dataset by computing the mean and standard deviation of purchase time gaps per category. A category is included if at least five users have repeated purchases in that category\footnote{The sampled dataset used for experimentation is a subset of production data and does not reflect the full platform scale.}.

\textbf{Implementation Details.}

We use LLAMA-3-8B-Instruct\footnote{\url{https://huggingface.co/NousResearch/Meta-Llama-3-8B-Instruct}} as the backbone and fine-tune it with LoRA on a single NVIDIA A10G (24GB). Training is conducted in float16 with batch size 1 and gradient accumulation of 64. We set LoRA hyperparameters to rank $r=64$, $\alpha=16$, and dropout $0.1$. Both fine-tuning stages share the same LoRA layers. To reduce memory usage, we apply 4-bit quantization.

\begin{table}[t]
\centering
\resizebox{1\columnwidth}{!}
{
\begin{tabular}{c|c||c|c}
\hline
\multicolumn{2}{c||}{\textbf{Conversion}} & \multicolumn{2}{c}{\textbf{Aisle}}  \\ \hline
Users          & 80,659  & Users &     206209         \\
Product Categories  &    4,904    & Aisles &  134   \\
Queries   &    1,563,868  & Product Names & 49,688  \\
Searches & 25,166,299 & Orders & 32,434,489\\
Conversion &  15,175,567 & Distinct Orders & 32,14,874 \\
Non-conversion & 13,295,855 & - & - \\ 

\hline
\end{tabular}
}
\caption{Statistics of the \emph{Conversion} and \emph{Aisle} dataset.}
\label{table:dataset}
\end{table}

\begin{table*}[t]
\centering

\resizebox{1\textwidth}{!}{
\begin{tabular}{l|c|c|c|c|c|c|c|c|c|c|c|c|c|c|c}
\hline
Metric\textbackslash{}Model 
& Glove & Cross Encoder & Two Tower & SASRec & LightGCN & ColBERT & CausalBERT 
& P5 & TIGER & SEATER & LLAMA3 & LLAMA3-ID 
& RecICL & A-LLMRec & \ours \\ \hline

Precision@5     
& 0.086 & 0.063 & 0.100 & 0.023 & 0.018 & 0.048 & 0.054 
& 0.023 & 0.082 & 0.017 & 0.075 & 0.049 
& 0.031 & 0.060 & \textbf{0.142} \\

Recall@5        
& 0.104 & 0.101 & 0.151 & 0.120 & 0.011 & 0.044 & 0.053 
& 0.018 & 0.059 & 0.010 & 0.120 & 0.049 
& 0.026 & 0.042 & \textbf{0.186} \\

F1@5            
& 0.079 & 0.067 & 0.099 & 0.037 & 0.010 & 0.037 & 0.043 
& 0.015 & 0.060 & 0.012 & 0.076 & 0.042 
& 0.024 & 0.041 & \textbf{0.126} \\ \hline

Precision@10    
& 0.060 & 0.048 & 0.077 & 0.018 & 0.018 & 0.033 & 0.041 
& 0.019 & 0.078 & 0.020 & 0.047 & 0.039 
& 0.020 & 0.036 & \textbf{0.094} \\

Recall@10       
& 0.139 & 0.161 & 0.182 & 0.190 & 0.009 & 0.061 & 0.073 
& 0.029 & 0.101 & 0.026 & 0.148 & 0.068 
& 0.033 & 0.049 & \textbf{0.231} \\

F1@10           
& 0.069 & 0.064 & 0.101 & 0.035 & 0.012 & 0.036 & 0.043 
& 0.019 & 0.079 & 0.020 & 0.059 & 0.042 
& 0.021 & 0.036 & \textbf{0.111} \\ \hline

Precision@20    
& 0.040 & 0.033 & 0.037 & 0.013 & 0.019 & 0.025 & 0.029 
& 0.013 & 0.068 & 0.037 & 0.030 & 0.029 
& 0.013 & 0.022 & \textbf{0.069} \\

Recall@20       
& 0.164 & 0.178 & 0.217 & 0.247 & 0.013 & 0.083 & 0.101 
& 0.040 & 0.152 & 0.088 & 0.171 & 0.105 
& 0.043 & 0.061 & \textbf{0.284} \\

F1@20           
& 0.056 & 0.047 & 0.056 & 0.025 & 0.007 & 0.033 & 0.040 
& 0.017 & 0.086 & 0.050 & 0.048 & 0.040 
& 0.018 & 0.028 & \textbf{0.097} \\

\hline
\end{tabular}
}
\caption{Evaluation on the \emph{Conversion} dataset.}
\label{table:conversionRes}
\end{table*}

\subsection{Baseline Models}\label{baselineMain}

We compare \ours\ with traditional and LLM-based baselines, including GLOVE~\cite{pennington2014glove}, Cross Encoder~\cite{reimers2019sentence}, Two Tower~\cite{huang2013learning}, SASRec~\cite{kang2018self}, LightGCN~\cite{he2020lightgcn}, ColBERT~\cite{khattab2020colbert}, CausalBERT~\cite{li2021causalbert}, P5~\cite{geng2022recommendation}, TIGER~\cite{rajput2024recommender}, SEATER~\cite{si2023generative}, LLAMA3~\cite{dubey2024llama}, LLAMA3-ID, RecICL~\cite{bao2024real}, and A-LLMRec~\cite{kim2024large}. Details are provided in Appendix~\ref{baseline}.

\subsection{Product Category Recommendation with Context Query}\label{exp1}

\noindent \textbf{Experiment Design and Metrics.}
We evaluate \ours\ on the \textbf{Conversion} dataset for category-level recommendation with contextual queries. Given a user's historical conversions and current session queries, the model predicts the top-$k$ product categories likely to convert. Both predictions and ground truth are sets of categories (ranking is not considered). We report the 5-run mean and standard deviation of Precision@K, Recall@K, and F1@K in Table~\ref{table:conversionRes}.

\noindent \textbf{Results and Discussion.}
\ours\ consistently outperforms all baselines across metrics, demonstrating strong modeling of both cyclical purchasing behavior and context-driven intent. Traditional retrieval models (GloVe, Cross Encoder, Two-Tower, ColBERT, CausalBERT) achieve moderate recall but lower precision due to the absence of structured decoding. Graph- and sequence-based methods (LightGCN, SASRec) struggle to align query semantics with category prediction. Semantic-ID approaches (TIGER, SEATER) underperform, likely due to limited supervision for ID construction.

Among LLM-based baselines, P5 and LLAMA3 show competitive recall but lack task-specific constraints. For \textbf{RecICL}, we adapt the original binary prediction setup to generate category lists, which deviates from its intended design and may limit performance. Similarly, \textbf{A-LLMRec} relies on rich item textual features for user representation learning; due to limited descriptive information in our dataset, we simplify its encoder, potentially restricting its effectiveness. Despite these adjustments, \ours\ maintains a substantial margin over all baselines, highlighting the benefit of explicit rebuying modeling and constrained generation.

\begin{table}[t]
\centering
\setlength{\tabcolsep}{3pt}
\resizebox{0.7\columnwidth}{!}{
\begin{tabular}{l|cccc|c}
\hline
Metric & AS1 & AS2 & AS3 & AS4 & \ours \\ \hline
P@5   & 0.074 & 0.133 & 0.108 & 0.000 & \textbf{0.138} \\
R@5   & 0.122 & 0.175 & 0.162 & 0.001 & \textbf{0.187} \\
F1@5  & 0.076 & 0.126 & 0.108 & 0.000 & \textbf{0.132} \\ \hline
P@10  & 0.046 & 0.087 & 0.067 & 0.000 & \textbf{0.094} \\
R@10  & 0.152 & 0.206 & 0.185 & 0.000 & \textbf{0.227} \\
F1@10 & 0.060 & 0.103 & 0.084 & 0.000 & \textbf{0.112} \\ \hline
P@20  & 0.030 & 0.061 & 0.044 & 0.000 & \textbf{0.069} \\
R@20  & 0.172 & 0.252 & 0.221 & 0.001 & \textbf{0.272} \\
F1@20 & 0.046 & 0.088 & 0.068 & 0.000 & \textbf{0.099} \\
\hline
\end{tabular}
}
\caption{Ablation study results.}
\label{table:ablation}
\end{table}

\subsection{Ablation Study}

In this section, we conduct an ablation study to assess the effectiveness of each proposed component of \ours on the \textbf{Conversion} dataset. The evaluation metrics remain consistent with previous experiments. The ablation includes the following: the vanilla LLAMA3 instruct model (AS1), \ours without Rebuying Cycle Information Injection (AS2), without Relational Information Fine-Tuning (AS3), and without Constrained Generation with Trie Tree (AS4). The experiment results are shown in Table~\ref{table:ablation}.

The analysis demonstrates that each component significantly enhances \ours’ performance. Notably, AS4 shows that the trie tree and beam search are crucial for generating valid tokens from the predefined category set, reinforcing the importance of Constrained Output Generation. These results underscore the necessity of each fine-tuning step in achieving optimal performance in the grocery recommendation task.

\subsection{Performance Comparison in a Production Environment} ~\label{prodExp}

We further evaluate \ours\ in a real-world production setting on a restocking task, which predicts the product categories a user is likely to purchase in the following week. This task requires jointly modeling user queries, historical conversions, cyclical repurchasing patterns, and inter-category relationships under practical system constraints.

\begin{figure}[h]
    \centering
    \includegraphics[width=1.0\columnwidth]{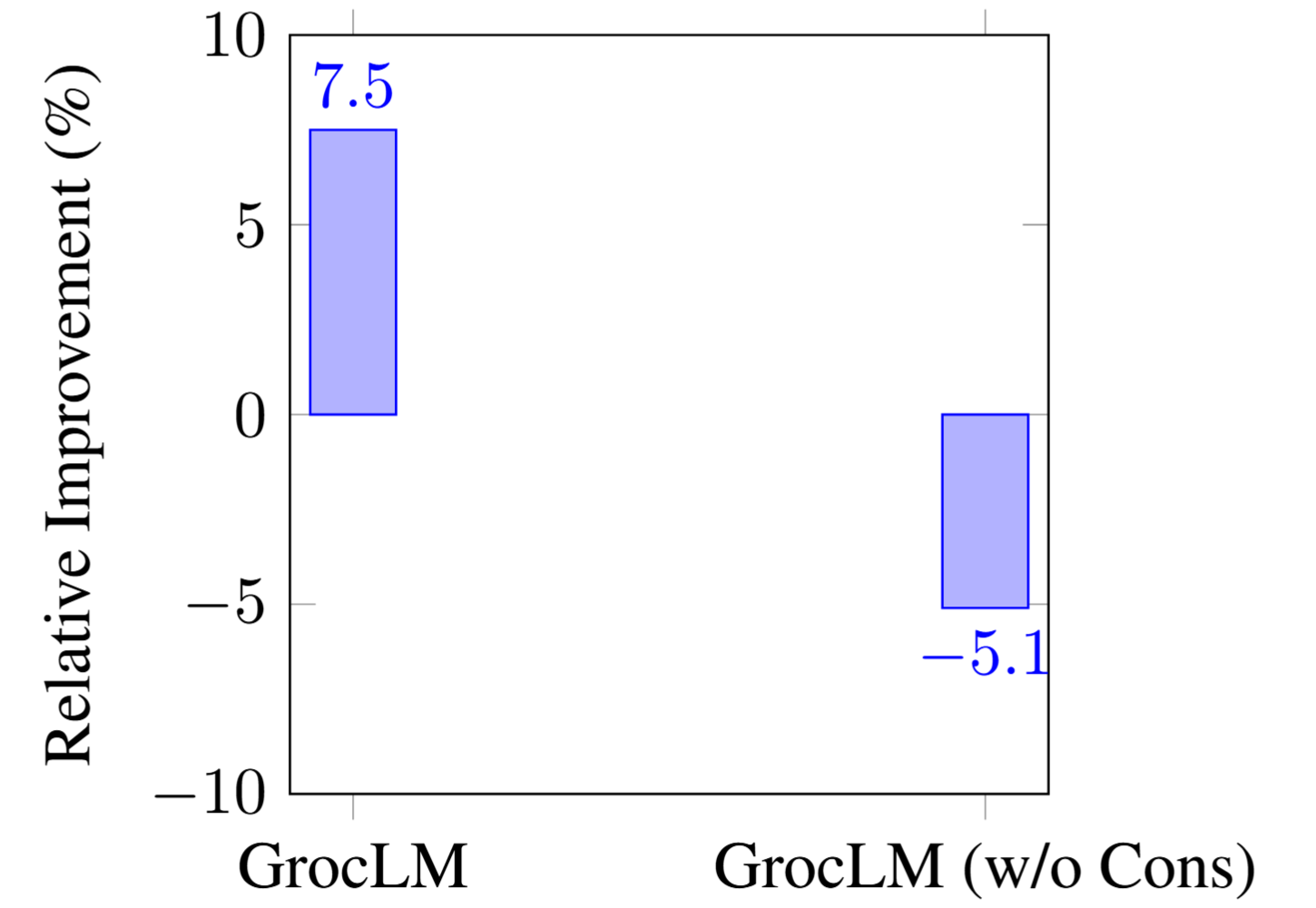}
    \caption{Relative improvement of \ours over the production model. (w/o Cons) refers to the variant without constrained generation.}
    \label{fig:relative_improvement}
\end{figure}

We compare \ours\ with the deployed production model. In this setup, candidate generation is restricted to each user’s previously purchased categories, enforced through dynamically loaded user-specific trie trees during decoding. As shown in Figure~\ref{fig:relative_improvement}, \ours\ achieves a 7.5\% relative improvement in \emph{cart-adds per impression} over the production model. In contrast, removing constrained generation and allowing unconstrained decoding leads to a 5.1\% performance regression, mainly due to duplicated or hallucinated category outputs.

Although trie-based masking introduces minor computational overhead, real-time latency remains acceptable for large-scale commercial deployment. In practice, \ours\ generates all categories jointly in $\sim$0.3s, while traditional methods require $\sim$0.5s per category, highlighting a substantial efficiency advantage. To further optimize latency, we are developing an event-driven system that precomputes recommendations via Kafka streams, enabling near real-time responses (sub-second latency) while preserving production constraints.

\begin{figure*}[h]
    \centering
    \includegraphics[width=0.95\textwidth]{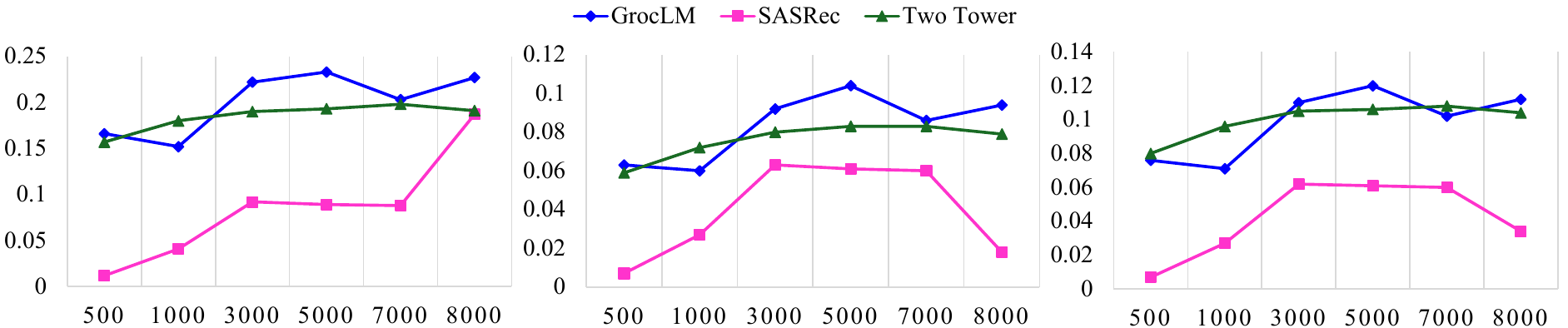}
    \caption{Data size sensitivity analysis. From Left to right: Recall@10, Precision@10, and F1@10.}
    \label{fig:sensitivity}
\end{figure*}

\subsection{Effect of Cycle Injection Strategy}
\label{sec:icl_rebuy}

To evaluate the role of Stage-1 cycle injection, we compare \ours\ with an in-context learning (ICL) variant where rebuying statistics are appended directly to the prompt in the form:

\textit{Rebuying stats: [<category, mean, std>]}.

Both approaches use identical inputs, differing only in whether cyclical information is encoded through parameter adaptation (LoRA) or provided via prompts. Compared to LoRA-based parameterization, the ICL formulation leads to longer prompts and less effective utilization of cyclical signals.

\begin{table}[t]
\centering
\setlength{\tabcolsep}{3pt}
\resizebox{1\columnwidth}{!}{
\begin{tabular}{l|ccc|ccc|ccc}
\hline
Model 
& P@5 & R@5 & F1@5 
& P@10 & R@10 & F1@10 
& P@20 & R@20 & F1@20 \\ \hline

ICL Only 
& 0.010 & 0.008 & 0.008
& 0.006 & 0.012 & 0.007
& 0.004 & 0.016 & 0.006 \\

\ours 
& \textbf{0.142} & \textbf{0.186} & \textbf{0.126}
& \textbf{0.094} & \textbf{0.231} & \textbf{0.111}
& \textbf{0.069} & \textbf{0.284} & \textbf{0.097} \\

\hline
\end{tabular}
}
\caption{Comparison between in-context learning (ICL) and LoRA-based cycle injection.}
\label{table:icl_ablation}
\end{table}

As shown in Table~\ref{table:icl_ablation}, the ICL-only variant performs substantially worse across all metrics. This demonstrates that directly providing statistical signals via prompts is insufficient, while LoRA-based parameter adaptation enables the model to internalize cyclical purchasing patterns more effectively. This also explains the robustness of \ours\ to prompt variations, as key information is encoded in model parameters rather than relying on prompt design.

\section{Data Size Sensitivity Analysis}

In this section, we perform a sensitivity analysis on the training data size to evaluate the impact of varying dataset sizes on the performance of different models with the \textbf{Conversion} dataset. Specifically, we analyze the models' performance in terms of Recall@10, Precision@10, and F1@10 across dataset sizes of 500, 1,000, 3,000, 5,000, 7,000, and 8,000 users from the original training dataset. We also fixed the testing data size to match the original testing dataset. We chose Two Tower and SASRec as baselines since they are widely used in the industry. We present the results in Figure~\ref{fig:sensitivity}.

As shown in the Figure, our model outperforms others as the dataset size increases, particularly after the dataset exceeds 3000 users. For smaller datasets (500–1,000 users), Two Tower shows marginally better results in precision and recall, likely due to its simpler architecture that performs well with limited data. However, as dataset sizes grow, both SASRec and Two Tower plateau around the 5,000-user mark. For SASRec, although we see an uptrend in recall with the increase of users, its precision drops significantly. In contrast, our model continues to improve, especially in larger datasets of 7,000 and 8,000 users, demonstrating its ability to capture more relevant items and refine predictions as more data becomes available. This trend is particularly evident in the F1@10 results, where our model maintains a better balance between recall and precision, outperforming the others as the dataset size increases. While Two Tower shows steady performance, it fails to achieve the same level of improvement as our model when the dataset grows.

%% file: inputs/5-conclusion.tex
\section{Conclusion}

We propose \ours, a generative language model for grocery category recommendation. By modeling user behavior at the category level and conditioning on historical purchases and queries, \ours\ captures cyclical patterns, diverse intent, and constrained outputs. Experiments show that \ours\ outperforms strong baselines, demonstrating effectiveness for real-world systems. Future work will explore low-resource personalization and broader category control.

%% file: inputs/7-supp.tex

\begin{table*}[]
\centering

\resizebox{1\textwidth}{!}{
\begin{tabular}{l|c|c|c|c|c|c|c|c|c|c|c|c|c}
\hline
Metric\textbackslash{}Model & Glove & Cross Encoder & Two Tower & SASRec & LightGCN & ColBERT & CausalBERT & P5 & TIGER & SEATER & LLAMA3 & LLAMA3-ID & \ours \\ \hline
Precision@5    & 0.046 & 0.048 & 0.111 & 0.091 & 0.007 & 0.086 & 0.125 & 0.002 & 0.021 & 0.051 & \textbf{0.137} & 0.000 & 0.128 \\
Recall@5       & 0.048 & 0.051 & 0.132 & 0.128 & 0.001 & 0.113 & 0.142 & 0.002 & 0.010 & 0.024 & 0.133 & 0.001 & \textbf{0.152} \\
F1@5           & 0.043 & 0.045 & 0.108 & 0.093 & 0.001 & 0.080 & 0.107 & 0.002 & 0.013 & 0.032 & 0.127 & 0.001 & \textbf{0.128} \\ \hline
Precision@10   & 0.051 & 0.045 & 0.085 & 0.065 & 0.017 & 0.092 & 0.103 & 0.003 & 0.024 & 0.055 & 0.085 & 0.055 & \textbf{0.118} \\
Recall@10      & 0.104 & 0.093 & 0.194 & 0.170 & 0.011 & 0.216 & 0.225 & 0.007 & 0.022 & 0.052 & 0.184 & 0.124 & \textbf{0.231} \\
F1@10          & 0.064 & 0.056 & 0.109 & 0.065 & 0.012 & 0.108 & 0.135 & 0.004 & 0.023 & 0.053 & 0.112 & 0.070 & \textbf{0.147} \\
\hline
\end{tabular}
}
\caption{Evaluation on the \emph{Aisle} dataset.}
\label{table:aisleRes}
\end{table*}

\section{Aisle Recommendation with In-basket Items}\label{exp2}
\noindent \textbf{Experiment Design and Evaluation Metrics.} In this task, we use the \textbf{Aisle} dataset to simulate a common grocery scenario where a user has already added items to their basket during a shopping session. The objective is to predict additional aisles the user is likely to visit based on the items already in the basket. In this context, product names are used to predict aisles, which is analogous to predicting categories from queries in the previous tasks. This setup allows us to assess the model's ability to leverage both direct relationships between in-basket items and broader historical patterns to make accurate aisle predictions.

To simulate the in-basket scenario, we randomly sample 50 percent of the products from each user's basket as input to the model and use the corresponding aisles of the remaining products as the ground truth. Historical aisle information is also provided, similar to the previous task. Given \textit{the smaller output space} of the \textbf{Aisle} dataset, we evaluate the model's performance with 5-run using the mean and standard deviations of Precision@K, Recall@K, and F1@K with $K$ set at 5 and 10. 

\noindent \textbf{Experiment Result Analysis and Discussion.}
Table~\ref{table:aisleRes} shows that \ours consistently outperforms all baselines in precision, recall, and F1, confirming its strength in modeling relationships between in-basket items and their corresponding aisles. While Two Tower and LLAMA3 perform relatively well, they still trail behind \ours, highlighting the benefit of incorporating structured decoding and historical context. ColBERT and CausalBERT achieve competitive results but are limited by their reliance on shallow token-level similarity. LightGCN also performs poorly, facing the same limitations observed in the previous experiment. P5 underperforms significantly, struggling to map in-basket items to aisles in this constrained setting.

\begin{figure*}
    \centering
    \includegraphics[width=0.95\textwidth]{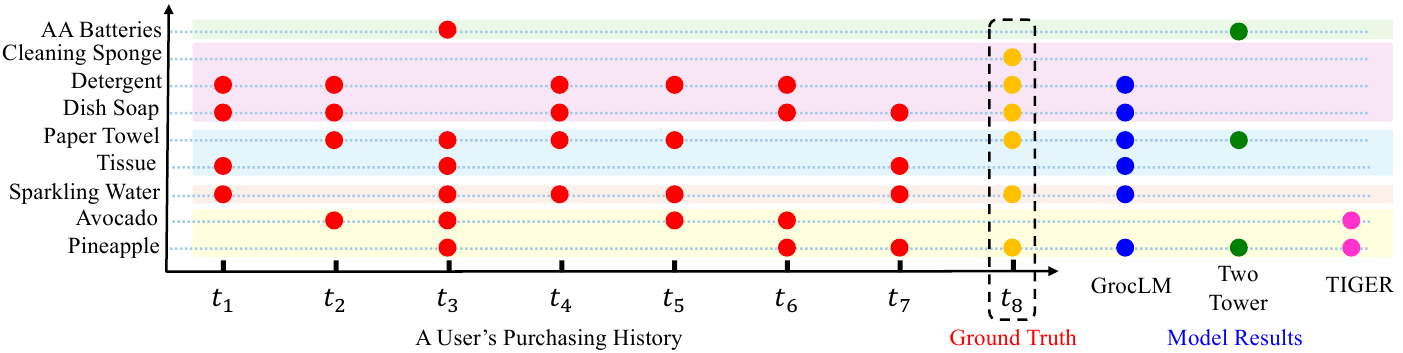}

    \caption{Illustration of the case study with a user's buying history.}
    \label{fig:caseStudy}

\end{figure*}

\section{Case Study}

\noindent \textbf{User Buying History.} To evaluate the effectiveness of our model in learning from historical purchasing patterns and predicting product categories accurately, we conducted a detailed case study, illustrated in Figure~\ref{fig:caseStudy}. The figure presents the anonymized buying history of a single user, that spans 8 times in total. The first 7 times, denoted as $t_1$ to $t_7$, represent the user's historical purchases, while the 8th serves as the ground truth and the prediction target. The vertical axis represents the product categories purchased over time. We applied the same background color to product categories that are alike to highlight related buying patterns. 

In addition to showcasing the predictions made by \ours, we compare the predictions of \ours against those of the Two Tower model, a widely-used baseline in industry, and TIGER, the recent semantic ID-based model. This comparison helps to assess each model's ability to capture and predict the user's purchasing behavior.

Our case study reveals that certain product categories exhibit consistent purchasing patterns several times, while others are more sporadic. \ours demonstrates its ability to accurately capture both types of patterns. For example, it correctly predicted the purchase of ``\textit{Sparkling Water},'' following a recurring trend of being bought three times with a one-time gap. Additionally, \ours successfully captured the co-occurrence of similar categories, such as ``\textit{Detergent}'' and ``\textit{Dish Soap},'' which also appear regularly in the user's buying history. In contrast, the Two Tower model misaligned, predicting ``\textit{AA Batteries},'' a product that is uncommon in the user's purchasing habits. Similarly, TIGER falsely predicted ``\textit{Avocado},'' a product frequently purchased by the user, but failed to capture its true skipping pattern. This discrepancy underscores \ours's strength in leveraging temporal data to accurately identify recurring categories. Furthermore, we observed that ``\textit{Cleaning Sponge}'' appears in the ground truth, but none of the models—including \ours—predicted it. This highlights the challenge of forecasting irregular or one-off purchases, which remains an area for improvement across all models. Despite these occasional missed predictions, \ours consistently shows a stronger grasp of recurring product relationships compared to the other models.

\begin{figure*}
    \centering
    \includegraphics[width=0.8\textwidth]{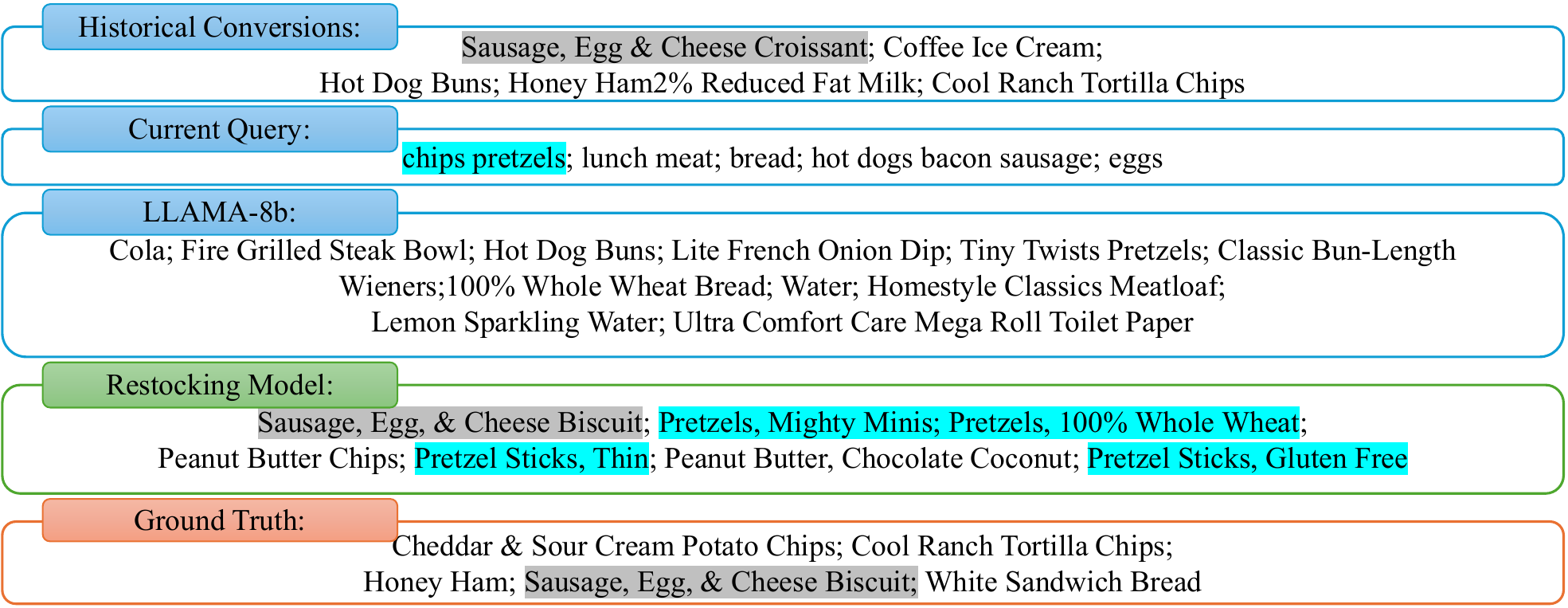}

    \caption{Illustration of the case study of the rebuying cycle.}
    \label{fig:newCaseStudy}

\end{figure*}

\noindent \textbf{Rebuying Cycle.} In this case study, we compare the vanilla LLAMA3-8b model with our Rebuying-Cycle–injected model to evaluate the effectiveness of the restocking mechanism. Both models were tested using the same prompting and evaluation strategy as the full model, with their predicted categories presented in Figure~\ref{fig:newCaseStudy}. The injected model clearly demonstrates its ability to leverage the user’s historical purchasing trends, resulting in more targeted and personalized predictions. For example, it successfully identifies recurring items like sausage-and-egg breakfast sandwiches (highlighted in gray). Additionally, the injected model responds effectively to the query, generating relevant recommendations (highlighted in blue). In contrast, the LLAMA-8b model produces a broader set of predictions that, while occasionally aligning with the user’s categories, often lack the precision needed to reflect true historical favorite, and  struggles to generate categories that belongs to the predefined category set.

In summary, the injection stage successfully inject the rebuying cycle into the model. However, the injection model does exhibit some overfitting tendencies, such as returning multiple closely related items rather than offering a more diverse set of suggestions, indicating future improvement directions.

\begin{table}[]
\centering

\vspace{-0.15in}
\resizebox{0.95\columnwidth}{!}{
\begin{tabular}{l|cccc|c}
\hline
Metric\textbackslash{}Model & AS1 & AS2 & AS3 & AS4 & \ours  \\ \hline
Acc: both       & 0.003 & 0.128 & 0.085 & 0.000 & \textbf{0.144}\\
Acc: board      & 0.002 & 0.115 & 0.080 & 0.000 & \textbf{0.132}\\
Acc: narrow     & 0.009 & 0.224 & 0.131 & 0.000 & \textbf{0.252}\\
\hline
\end{tabular}
}
\caption{Pairwise evaluation.}
\label{table:broadAndNarrow}
\end{table}
\section{Pairwise Evaluation of Query Intent}

Testing how the model handles broad and narrow queries is crucial for assessing its ability to generalize across varying levels of specificity, which is essential for real-world applications where user queries can range from highly general to very specific. In this experiment, we evaluate how \ours responds to broad queries like ``fruits'' and ``meats'' versus narrow queries such as ``Coca-Cola.'' Intuitively, narrow queries should be easier to handle due to their specificity, while broad queries are more challenging due to their vagueness.

For this task, we use the \textbf{Conversion} dataset and first classify each user query intent as broad or narrow using the LLAMA3 8B Instruction model as a data pre-processing step. We input the product category and constrain the output to be either broad or narrow. We also exclude queries that did not result in a conversion. Each query is used to predict the paired conversion product category, generating one output candidate per query. Accuracy is used as the evaluation metric to measure the model’s ability to correctly predict the intended category.

This experiment is conducted in three parts: evaluating broad queries only, narrow queries only, and a combination of both. We experiment on all model variants from the ablation study. The results are presented in Table~\ref{table:broadAndNarrow}. The results align with our expectation that narrow queries are easier for the model to handle, resulting in more accurate predictions. The vanilla LLAMA3 model (AS1) performs poorly, likely due to its lack of fine-tuning on the subset of product categories and limited understanding of their relationships. As the level of fine-tuning increases, the model's performance improves, with the best results achieved by the full \ours. This demonstrates the effectiveness of the additional tuning steps in enhancing the model’s ability to process different query intents accurately.

\section{Baseline Models} \label{baseline}

The details of the chosen baseline models are:

\begin{itemize}
    \item \textbf{GLOVE}~\cite{pennington2014glove} is an unsupervised learning algorithm for obtaining vector representations of words. GLOVE captures semantic relationships between words by training on aggregated global word-word co-occurrence statistics from a corpus.
    
    \item \textbf{Cross Encoder}~\cite{reimers2019sentence} processes pairs of input texts together and generates a single embedding for the combined input, allowing detailed interaction between the inputs.
    
    \item \textbf{Two Tower}~\cite{huang2013learning} independently encodes the query and candidate items with separate encoders and compares the generated embeddings using a similarity measure. It is efficient for large-scale retrieval tasks as it allows for the pre-computation and indexing of candidate item embeddings.

    \item \textbf{LightGCN}~\cite{he2020lightgcn} is a graph-based collaborative filtering model that simplifies traditional GCNs by removing feature transformation and nonlinear activation, focusing purely on neighborhood aggregation over a user-item interaction graph.

    \item \textbf{ColBERT}~\cite{khattab2020colbert} is a late interaction retrieval model that encodes queries and documents into contextualized token-level embeddings and performs similarity computation via maximum-similarity matching across tokens, allowing efficient and accurate retrieval.

    \item \textbf{CausalBERT}~\cite{li2021causalbert} integrates causal inference with language modeling by estimating treatment effects from observational data. In retrieval, it leverages token-level representations with causality-aware modeling to improve recommendation accuracy.
    
    \item \textbf{SASRec}~\cite{kang2018self} uses self-attention mechanisms to model user behavior sequences, capturing temporal dynamics and contextual dependencies.
    
    \item \textbf{P5}~\cite{geng2022recommendation} is a unified model that uses a text-to-text paradigm for various recommendation tasks. By converting all data into natural language sequences, P5 leverages deep semantics for personalized recommendations, enabling zero-shot and few-shot predictions across different tasks.
    
    \item \textbf{TIGER}~\cite{rajput2024recommender} encodes item-related information with a Residual-Quantized Variational Autoencoder (RQ-VAE)~\cite{lee2022autoregressive} and generates a corresponding semantic ID. With the semantic ID replacing the original items, it utilizes a variant of  T5~\cite{raffel2020exploring} to provide sequential recommendations.
    
    \item \textbf{SEATER}~\cite{si2023generative} not only constructs semantic IDs for each item, but also further distinguishes items with a distinguishing loss and a ranking loss. It also utilizes a variant of the T5~\cite{raffel2020exploring} for recommendations.
    
    \item \textbf{LLAMA3}~\cite{dubey2024llama} is a large language model that leverages extensive pre-training on diverse datasets to perform a wide range of generative tasks, demonstrating the capabilities of state-of-the-art language models in understanding and generating coherent text.
    
    \item \textbf{LLAMA3-ID} follows a similar training and evaluation procedure as \ours, but instead of generating natural text, this model directly outputs the IDs of the items.
    
    \item \textbf{RecICL}~\cite{bao2024real} is an in-context learning framework for recommendation that performs prediction by constructing task-specific prompts from historical interactions.
    
    \item \textbf{A-LLMRec}~\cite{kim2024large} combines encoder-based user representation learning with LLM prompting to perform history-to-item recommendation.
\end{itemize}


\begin{table*}[h]
\centering

\resizebox{0.75\textwidth}{!}{
\begin{tabular}{l|l}
\hline
\textbf{User Query} & \textbf{Predicted Product Categories} \\ \hline
cookie baking chocolate chips flour & Chocolate Chips, Flour \\
movie night chips soda              & Potato Chips, Popcorn, Soft Drinks \\
breakfast eggs bread milk           & Eggs, Bread, Milk \\
kitchen cleaner                     & All-Purpose Cleaner, Disinfectant Spray, Scrub Sponges \\
pretzel snack                       & Pretzel Bites, Pretzel Rods, Pretzel Sticks, Pretzel Rings \\
<No User Query>                                  & Baking Supplies \\
\hline
\end{tabular}
}
\caption{Cold-start predictions from GrocLM using only user query keywords.}
\label{table:coldstart}
\end{table*}

\section{Cold-Start and Sparse User Behavior}\label{coldstart}

\begin{mybox}
$\bullet$ \textbf{Task Description:} ``\textit{Task: Recommend product categories given a new user's search query.}''

$\bullet$ \textbf{Prompt Structure:} ``\textit{The following are examples for input and output structure:}''

\quad Example 1: Input: \{snacks\}. Output: \{Vegetable Chips, Crackers, Snack Packs, Seaweed Chips, Packaged Cookies\}.

\quad Example 2: Input: \{fruit\}. Output: \{Strawberries, Blackberries, Watermelons, Green Grapes, Bananas\}.

$\bullet$ \textbf{Generation Instruction:} ``\textit{Given the input below, generate output following the format above.}''

$\bullet$ \textbf{Query Template:} Input: \{\textit{<user query>}\}. Output:\{
\end{mybox}

To evaluate GrocLM's performance in cold-start scenarios, we conducted a controlled experiment using a template-style prompt that includes only the user's current search query without any prior conversion history. This setup simulates a realistic case where a new or infrequent user initiates a session with no behavioral history. Additionally, to simulate a complete cold-start condition, we also test the model with an empty query input, removing all contextual signals. The results of these experiments are shown in Table~\ref{table:coldstart}. We observe that even in the absence of any user history, GrocLM is able to interpret free-form keyword queries and generate relevant product categories from the predefined vocabulary set, demonstrating its ability to generalize from language cues. However, when both history and query input are absent, the model produces a static output:\textit{Baking Supplies}, likely reflecting common patterns seen during training. This highlights a key limitation in the current framework and motivates future work to enhance cold-start robustness. Specifically, we plan to incorporate auxiliary user-related signals such as demographics, time-of-day, session metadata, or globally trending items to better personalize recommendations in zero-input scenarios and improve model flexibility under sparse conditions.

\section{Discussion on Generalization to Other Domains}\label{general}

While \ours\ is specifically designed for grocery category recommendation, its architectural components are generalizable to other e-commerce domains that exhibit structured taxonomies and sequential purchase behaviors. In particular:
\begin{itemize}
\item The two-stage prompt tuning strategy can be adapted using alternative domain-specific priors (e.g., seasonal trends in fashion or replenishment cycles in pet supplies).
\item The trie-constrained decoding mechanism is applicable wherever a fixed, valid label space exists.
\item The use of precomputed rebuying statistics or similar global signals can help bootstrap recommendations in sparse settings across domains.
\end{itemize}

We leave domain transfer experiments to future work, but we believe the modeling framework presented in this paper lays a foundation for extensibility to other structured recommendation tasks.

\section{Error Analysis through User Example} \label{errorAnalysis}

Based on case study results in Figure~\ref{fig:caseStudy} and Figure~\ref{fig:newCaseStudy}, we identify two representative error types exhibited by our model. First, GrocLM occasionally struggles to capture implicit product relationships. For example, it successfully predicts Detergent and Dish Soap but overlooks the complementary Cleaning Sponge, revealing a limitation in its in-context relational reasoning. Second, the model may overfit to frequent patterns learned from rebuying-cycle supervision, leading to redundant or overly similar predictions—such as generating multiple variants of Pretzels—which reduces overall output diversity. These observations highlight opportunities to enhance relational reasoning and improve the balance between consistency and coverage in future model iterations.